\documentclass{article}

\usepackage{microtype}
\usepackage{graphicx}
\usepackage{booktabs} 

\usepackage{hyperref}

\usepackage[accepted]{icml2025}

\usepackage{amsmath}
\usepackage{amssymb}
\usepackage{mathtools}
\usepackage{amsthm}

\usepackage[capitalize,noabbrev]{cleveref}

\usepackage{amsmath,amsfonts,bm}

\def\eqref#1{equation~\ref{#1}}

\def\1{\bm{1}}

\def\vw{{\bm{w}}}

\DeclareMathAlphabet{\mathsfit}{\encodingdefault}{\sfdefault}{m}{sl}
\SetMathAlphabet{\mathsfit}{bold}{\encodingdefault}{\sfdefault}{bx}{n}

\DeclareMathOperator*{\argmax}{arg\,max}

\usepackage{dsfont}
\usepackage{bm}
\usepackage{bbm}
\usepackage{graphicx}
\usepackage{color}
\usepackage{multicol}
\usepackage{multirow}
\usepackage{wrapfig,lipsum,booktabs}
\usepackage{pgfplots}
\pgfplotsset{compat=1.18}
\usepackage{tikz}
\usepackage{xcolor}
\usepackage{colortbl}
\usepackage{xspace}
\usepackage{makecell}
\usepackage{soul}
\usepackage{amsmath,amsfonts,amssymb}
\usepackage{subcaption}
\usetikzlibrary{calc}
\usepgfplotslibrary{groupplots}
\usetikzlibrary{angles,quotes} 
\usetikzlibrary{shapes,arrows}
\usetikzlibrary{backgrounds}
\usetikzlibrary{matrix}
\usetikzlibrary{patterns}
\usetikzlibrary{angles,quotes,shapes,arrows,calc,chains,fit,positioning,shapes}
\usepackage{tikz-3dplot}
\usepackage{hyperref}
\usepackage{cleveref}
\usepackage{paralist}
\usepackage{cancel}
\usepackage{tabu}
\usepackage{rotating}
\usepackage{etoolbox}
\usepackage{adjustbox}
\usepackage{enumerate}
\usepackage{enumitem}
\setitemize{itemsep=0pt,topsep=0pt,parsep=0pt,partopsep=0pt}
\setenumerate{itemsep=0pt,topsep=0pt,parsep=0pt,partopsep=0pt}
\usepackage{pifont}
\usepackage{cancel}
\usepackage{lipsum}
\usepackage{listings,lstautogobble}
\usepackage{fancyvrb}
\usepackage{fvextra}
\usepackage{caption}
\usepackage{arydshln}

\newcommand{\model}[1]{\textsc{#1}\xspace}
\newcommand{\dataset}[1]{\texttt{#1}\xspace}

\newcommand{\gemma}{\model{Gemma-2-2b}}
\newcommand{\mistral}{\model{Mistral-7B-v0.3}}
\newcommand{\llama}{\model{Llama-3-8B}}

\newcommand{\random}{\model{Random}}
\newcommand{\longest}{\model{Longest}}
\newcommand{\ppl}{\model{Perplexity}}
\newcommand{\armorm}{\model{ArmoRM}}
\newcommand{\alpagasus}{\model{AlpaGasus}}
\newcommand{\deita}{\model{Deita}}
\newcommand{\ifd}{\model{SuperFilter}}
\newcommand{\instag}{\model{InsTag}}
\newcommand{\kmeans}{\model{Kmeans}}

\newcommand{\ours}{\model{GraphFilter}}

\newcommand{\mmlu}{\dataset{MMLU}}
\newcommand{\arc}{\dataset{ARC}}
\newcommand{\hellaswag}{\dataset{HellaSwag}}
\newcommand{\gsmk}{\dataset{GSM8K}}
\newcommand{\alpacaeval}{\dataset{AlpacaEval-2.0}}

\newcommand{\mtbench}{\dataset{MT-Bench}}

\newcommand{\avgall}{$\mu_{\textsc{all}}$\xspace}
\newcommand{\avgbench}{$\mu_{\textsc{bench}}$\xspace}
\newcommand{\avgllm}{$\mu_{\textsc{llm}}$\xspace}
\newcommand{\avgmt}{$\mu_{\textsc{mt}}$\xspace}

\newcommand{\cmark}{\ding{51}}%
\newcommand{\xmark}{\ding{55}}%
\newcommand{\checksymbol}{\cmark\xspace}
\newcommand{\crosssymbol}{\xmark\xspace}
\definecolor{tgreen}{HTML}{66c56c}
\definecolor{tyellow}{HTML}{FDDA0D}

\theoremstyle{plain}

\theoremstyle{definition}

\theoremstyle{remark}

\usepackage[textsize=tiny]{todonotes}

\icmltitlerunning{The Best of Both Worlds: Bridging Quality and Diversity in Data Selection with Bipartite Graph}

\begin{document}

\renewcommand{\tableautorefname}{Table}
\renewcommand{\sectionautorefname}{Section}
\renewcommand{\subsectionautorefname}{Section}
\renewcommand{\subsubsectionautorefname}{Section}
\renewcommand{\figureautorefname}{Figure}
\renewcommand{\equationautorefname}{Equation}
\newcommand{\algorithmautorefname}{Algorithm}
\newcommand{\linenoautorefname}{Line}

\twocolumn[
\icmltitle{The Best of Both Worlds: Bridging Quality and Diversity in\\Data Selection with Bipartite Graph}



\icmlsetsymbol{equal}{*}

\begin{icmlauthorlist}
\icmlauthor{Minghao Wu}{yyy}
\icmlauthor{Thuy-Trang Vu}{yyy}
\icmlauthor{Lizhen Qu}{yyy}
\icmlauthor{Gholamreza Haffari}{yyy}
\end{icmlauthorlist}

\icmlaffiliation{yyy}{Department of Data Science \& AI, Monash University, Melbourne, Australia}

\icmlcorrespondingauthor{Minghao Wu}{minghao.wu@monash.edu}

\icmlkeywords{Machine Learning, ICML}

\vskip 0.3in
]



\printAffiliationsAndNotice{}  

\begin{abstract}

The performance of large language models (LLMs) is strongly influenced by the quality and diversity of data used during supervised fine-tuning (SFT). However, current data selection methods often prioritize one aspect over the other, resulting in suboptimal training outcomes. To address this, we formulate data selection as a set cover problem and present \ours, a novel approach that balances both quality and diversity in data selection. \ours models the dataset as a bipartite graph connecting sentences to their constituent n-grams, then employs a priority function that combines quality and diversity metrics multiplicatively. \ours iteratively selects sentences with the highest priority, removes covered n-grams from the bipartite graph, and recomputes priorities to reflect the changing data landscape. We validate \ours using three model backbones across six widely-used benchmarks, demonstrating that it outperforms nine existing baselines in both model performance and computational efficiency. Further analysis shows that our design choices lead to more effective subset selection, underscores the value of instruction diversity, and provides insights into how quality and diversity interact with different subset sizes.

\end{abstract}

\section{Introduction}

Large language models (LLMs) have significantly advanced the field of natural language processing (NLP), enabling models to generate coherent and contextually relevant text across a variety of tasks \citep{DBLP:conf/nips/Ouyang0JAWMZASR22, DBLP:conf/iclr/SanhWRBSACSRDBX22,DBLP:journals/corr/abs-2303-08774, DBLP:journals/corr/abs-2302-13971,DBLP:journals/corr/abs-2307-09288,DBLP:journals/corr/abs-2312-11805,DBLP:journals/corr/abs-2403-08295,DBLP:journals/corr/abs-2407-10671}. Central to the success of these models is the quality and diversity of the data used during supervised fine-tuning (SFT). Fine-tuning on high-quality data ensures that the model learns accurate language patterns and responds appropriately to inputs \citep{wang-etal-2023-self-instruct,DBLP:conf/nips/ZhouLX0SMMEYYZG23}, while diversity in the data allows the model to generalize across different contexts and topics \citep{DBLP:journals/corr/abs-2303-09540,DBLP:conf/iclr/MaharanaYB24}. However, the vastness of available SFT data presents a challenge: selecting a subset of data that balances both quality and diversity to optimize model performance.

Recent methods for data selection often prioritize either quality or diversity, rarely achieving an optimal balance of both. Approaches that focus exclusively on quality may overlook the variety of language patterns necessary for effective generalization \citep{DBLP:journals/corr/abs-2309-04564,DBLP:journals/corr/abs-2405-20541}. Conversely, methods emphasizing diversity might include lower-quality data, which could negatively impact model performance \citep{DBLP:journals/corr/abs-2303-09540,DBLP:conf/iclr/LuY0LLTZZ24}. This focus can result in models that either overfit to specific data patterns or underperform due to the inclusion of irrelevant or poor-quality data. Hence, it is crucial to develop a data selection strategy that simultaneously maximizes both data quality and diversity for effective supervised fine-tuning.

In response to this challenge, we formulate data selection as a \textit{set cover problem} and propose a novel method, \ours, which models both \textit{diversity} and \textit{quality} in data selection. The set cover problem aims to select the smallest collection of subsets to cover every element in a given universal set \citep{DBLP:books/fm/GareyJ79}. To achieve this in data selection, \ours models the dataset as a bipartite graph, where sentences and n-grams are represented as two distinct sets of nodes, with edges indicating the presence of n-grams in sentences. This bipartite structure allows us to prioritize sentences that introduce unique n-grams, thereby maximizing the diversity of the selected subset. In addition to diversity, \ours incorporates quality into the selection process by re-ranking sentences based on a quality metric. To balance these two aspects, we employ a priority function that combines quality and diversity metrics multiplicatively. Concretely, we use \ifd \citep{li-etal-2024-superfiltering} as the quality metric. It measures the informativeness of a response by comparing its perplexity when conditioned on the instruction with its standalone perplexity. Moreover,  we leverage Term Frequency-Inverse Document Frequency (\model{TF-IDF}) scores for n-grams within sentences as the diversity metric.  The priority function multiplies these two measures, assigning higher priority to sentences that are both informative (high-quality) and contribute substantially to n-gram diversity. During selection, \ours iteratively chooses the sentence with the highest priority, updates the bipartite graph by removing covered n-grams, and recalculates priorities based on the updated graph. By balancing diversity and quality in this manner, \ours effectively build a subset of examples that is both high-quality and broadly representative of the entire dataset.

To demonstrate the effectiveness of \ours, we conducted extensive experiments, comparing \ours against nine baseline approaches using three model backbones across six widely-used benchmarks. Our empirical results indicate that \ours significantly outperforms recent state-of-the-art baselines and achieves notably better computational efficiency. Specifically, in terms of overall performance, \ours substantially outperforms all the baseline approaches across three model backbones, and is significantly more efficient than most baselines without requiring GPUs for computation. \ours outperforms the baselines by up to $+2.37$ for \gemma, $+3.02$ for \mistral, and $+3.38$ for \llama, and is significantly more efficient than these baselines without requiring GPUs for computation. Furthermore, we perform an in-depth analysis to validate the effectiveness of our design choices in \ours, examine the characteristics of the selected subsets and the importance of instruction diversity, and investigate the impacts of quality and diversity under various data scales.

In summary, the contributions of this work are threefold:
\begin{itemize} 
    \item We frame data selection as a set cover problem and introduce a novel approach, \ours, which leverages a bipartite graph structure to balance both diversity and quality. This bipartite graph connects sentences to their constituent n-grams, enabling the construction of a subset of examples that is not only high-quality but also broadly representative of the entire dataset (see \autoref{sec:method}). 
    
    \item Through experiments using three model backbones across six widely-used benchmarks, we demonstrate that our method, \ours, surpasses existing data selection strategies, achieving significantly better computational efficiency (see \autoref{sec:experiments}). 
    
    \item Our detailed analyses provide valuable insights into the design choices of \ours, the characteristics of the selected subset, and the importance of quality and diversity in relation to the subset sizes (see \autoref{sec:analysis}).
\end{itemize}

\section{Related Work}

\paragraph{Data Engineering for Large Language Models}
The success of recent large language models (LLMs) largely relies on the data used during their training process \citep{DBLP:journals/corr/abs-2303-10158}. State-of-the-art LLMs are generally trained on vast corpora \citep{DBLP:journals/corr/abs-2303-08774, gemmateam2024gemma2improvingopen, DBLP:journals/corr/abs-2407-21783}. A significant area of research focuses on curating high-quality corpora for pre-training these models \citep{DBLP:journals/jmlr/RaffelSRLNMZLL20, together2023redpajama, soldaini-etal-2024-dolma, DBLP:journals/corr/abs-2406-17557}. Furthermore, \citet{wang-etal-2023-self-instruct} demonstrate that LLMs are capable of synthesizing high-quality datasets for supervised fine-tuning, which leads to a surge of research on dataset synthesis \citep{DBLP:journals/corr/abs-2304-12244, DBLP:journals/corr/abs-2305-15011, DBLP:journals/corr/abs-2306-11644, ding-etal-2023-enhancing, DBLP:journals/corr/abs-2310-01377, wu-etal-2024-lamini, DBLP:journals/corr/abs-2406-10323, DBLP:journals/corr/abs-2406-08464}. These research efforts facilitate the synthesis of large-scale datasets containing billions of tokens for various purposes, resulting in a significant demand for selecting valuable subsets.

\paragraph{Data Selection}
Data selection strategies aim to identify the most informative data subsets for training or fine-tuning models by considering quality and diversity. Quality-focused approaches prioritize metrics like complexity, difficulty, or informativeness \citep{DBLP:journals/corr/abs-2309-04564,DBLP:conf/iclr/ChenLYWGYTS0HJ24,DBLP:conf/iclr/0131Z00H24,li-etal-2024-quantity,li-etal-2024-superfiltering}, but may neglect the range of language patterns needed for generalization. Conversely, diversity-focused methods capture a broad spectrum of linguistic patterns and contexts, potentially incorporating lower-quality data that could impair model performance \citep{DBLP:journals/corr/abs-2303-09540,DBLP:conf/iclr/LuY0LLTZZ24}.

\paragraph{Ours}
To overcome limitations in current data selection methods, we propose \ours, a novel approach that represents the dataset as a bipartite graph of sentences and their n-grams. By balancing quality and diversity with a priority function, our method improves model performance across various downstream tasks.

\section{Methodology}
\label{sec:method}

In this section, we first introduce the data selection problem for supervised fine-tuning in \autoref{sec:problem}. Subsequently, we describe the modeling of the dataset as a bipartite graph in \autoref{sec:graph}. Finally, we explain the re-ranking of the graph nodes using a priority function that integrates quality and diversity metrics in data selection in \autoref{sec:priority}.

\subsection{Data Selection Problem}
\label{sec:problem}

The data selection problem involves the challenge of identifying and selecting the most relevant and informative subset of supervised instances from a larger dataset to fine-tune large language models (LLMs). Formally, let $\mathcal{D} = \{(x_i, y_i)\}_{i=1}^N$ be the supervised fine-tuning (SFT) dataset, where $x_i$ represents the instruction and $y_i$ its corresponding response for the $i$-th training instance. Our aim is to select a subset $\mathcal{S}_{\pi}$ of size $k$ from $\mathcal{D}$, utilizing the data selection strategy $\pi$, where $k$ is the \textit{data selection budget}. The objective is to determine the optimal data selection strategy $\pi^*$ that is capable of selecting a subset $\mathcal{S}_{\pi}$ maximizing the performance of the fine-tuned LLM $f_{\theta}$ on the downstream tasks $\mathcal{D}_{\text{tst}}$. Therefore, the data selection problem can be formally formulated as:


\begin{equation}
\begin{aligned}
\pi^* &= \argmax_{\pi} \, \mathcal{R}\left( f_\theta; \mathcal{D}_{\text{tst}} \right) \text{, subject to } |\mathcal{S}_{\pi}| = k, \\ &\text{where } \theta = \text{FineTune}(\mathcal{F}, \mathcal{S}_{\pi}),
\label{eq:definition}
\end{aligned}
\end{equation}

where $\mathcal{S}_{\pi}$ is the subset of the training data selected by the strategy $\pi$, $\theta = \text{FineTune}(\mathcal{F}, \mathcal{S}_{\pi})$ denotes the parameters of the model backbone $\mathcal{F}$ after fine-tuning on the selected data subset $\mathcal{S}_{\pi}$, $f_\theta$ is the fine-tuned model with parameters $\theta$, and $\mathcal{R}\left( f_\theta; \mathcal{D}_{\text{tst}} \right)$ is the performance metric (e.g., accuracy) of the fine-tuned model $f_\theta$ evaluated on the test set $\mathcal{D}_{\text{tst}}$.

\begin{algorithm}[t]
\caption{\ours}
\label{alg:graphfilter}
\begin{algorithmic}[1]

\STATE {\bfseries Input:} $\mathcal{U}=\{u_i\}^{N}_{i=1}$, the set of sentence nodes; $\mathcal{V}=\{v_j\}^{M}_{j=1}$, the set of n-gram nodes; $\mathcal{E} \subseteq \mathcal{U} \times \mathcal{V}$, the set of edges between sentence nodes and n-gram nodes; $k$, the data selection budget; $\phi(u)$, the priority function for each $u \in \mathcal{U}$;
\STATE {\bfseries Output:} The selected subset $\mathcal{S}$;
\STATE $\mathcal{S}=\emptyset$\;
\WHILE{$|\mathcal{S}| < k \land \mathcal{U} \neq \emptyset$}
\STATE $u^* \leftarrow \argmax_{u \in \mathcal{U}} \phi(u)$\; \COMMENT{\texttt{Select the sentence with the highest priority}}
\STATE $\mathcal{V}_{u^*} \leftarrow \{ v \in \mathcal{V} \mid (u^*, v) \in \mathcal{E} \}$\; \COMMENT{\texttt{Find n-gram nodes connected to $u^*$}}
\STATE $\mathcal{S} \leftarrow \mathcal{S} \cup \{ u^* \}$\; \COMMENT{\texttt{Add $u^*$ to the selected set}}
\STATE $\mathcal{U} \leftarrow \mathcal{U} \setminus \{ u^* \}$\; \COMMENT{\texttt{Remove $u^*$ from the remaining sentences}}
\STATE $\mathcal{E} \leftarrow \mathcal{E} \setminus \{ (u^*, v) \mid v \in \mathcal{V}_{u^*} \}$\; \COMMENT{\texttt{Remove edges connected to $u^*$}}
\FORALL{$v \in \mathcal{V}_{u^*}$} 
\STATE $\mathcal{E} \leftarrow \mathcal{E} \setminus \{ (u, v) \mid u \in \mathcal{U} \}$\; \COMMENT{\texttt{Remove edges connecting to $v$}}

\ENDFOR
\ENDWHILE

\end{algorithmic}
\end{algorithm}





 \begin{figure*}[ht]
    \centering
    \begin{subfigure}[]{0.2\textwidth}
        \centering
        \begin{tikzpicture}[
            node distance = 1.5mm and 15mm,
              start chain = going below,
                 V/.style = {circle, draw, 
                             fill=#1, 
                             node contents={}},
                            ]
            \node (n14) [V=blue,on chain,label={left:$u_{1}$}];
            \node (n13) [V=blue,on chain,label={left:$u_{2}$}];
            \node (n12) [V=blue,on chain,label={left:$u_{3}$}];
            \node (n11) [V=blue,on chain,label={left:$u_{4}$}];
            \node (n10) [V=blue,on chain,label={left:$u_{5}$}];

            \node (n24) [V=tgreen,right=1cm of n14,label={right:$v_{1}$}];
            \node (n23) [V=tgreen,right=1cm of n13,label={right:$v_{2}$}];
            \node (n22) [V=tgreen,right=1cm of n12,label={right:$v_{3}$}];
            \node (n21) [V=tgreen,right=1cm of n11,label={right:$v_{4}$}];
            \node (n20) [V=tgreen,right=1cm of n10,label={right:$v_{5}$}];
            \node [blue,fit=(n14) (n10),label=above:$\mathcal{U}$] {};
            \node [green,fit=(n24) (n20),label=above:$\mathcal{V}$] {};
            \draw[-]
            (n14) edge (n24) (n14) edge (n23) (n14) edge (n22)
            (n13) edge (n23) (n13) edge (n22)
            (n12) edge (n24) (n12) edge (n21)
            (n11) edge (n21) (n11) edge (n20)
            (n10) edge (n24) (n10) edge (n20);
        
            \end{tikzpicture}
            \caption{Initial Graph}
            \label{fig:example_initial}
    \end{subfigure}
    \hspace{10pt}
    \begin{subfigure}[]{0.2\textwidth}
        \centering
        \begin{tikzpicture}[
            node distance = 1.5mm and 15mm,
              start chain = going below,
                 V/.style = {circle, draw, 
                             fill=#1, 
                             node contents={}},
                            ]
            \node (n14) [V=yellow,on chain,label={left:$u_{1}$}];
            \node (n13) [V=blue,on chain,label={left:$u_{2}$}];
            \node (n12) [V=blue,on chain,label={left:$u_{3}$}];
            \node (n11) [V=blue,on chain,label={left:$u_{4}$}];
            \node (n10) [V=blue,on chain,label={left:$u_{5}$}];

            \node (n24) [V=red,right=1cm of n14,label={right:$v_{1}$}];
            \node (n23) [V=red,right=1cm of n13,label={right:$v_{2}$}];
            \node (n22) [V=red,right=1cm of n12,label={right:$v_{3}$}];
            \node (n21) [V=tgreen,right=1cm of n11,label={right:$v_{4}$}];
            \node (n20) [V=tgreen,right=1cm of n10,label={right:$v_{5}$}];
            \node [blue,fit=(n14) (n10),label=above:$\mathcal{U}$] {};
            \node [green,fit=(n24) (n20),label=above:$\mathcal{V}$] {};

            \draw[-]
            (n14) edge (n24) (n14) edge (n23) (n14) edge (n22)
            (n13) edge (n23) (n13) edge (n22)
            (n12) edge (n24) (n12) edge (n21)
            (n11) edge (n21) (n11) edge (n20)
            (n10) edge (n24) (n10) edge (n20);

            \end{tikzpicture}
            \caption{Selection}
            \label{fig:example_selection}
    \end{subfigure}
    \hspace{10pt}
    \begin{subfigure}[]{0.2\textwidth}
        \centering
        \begin{tikzpicture}[
            node distance = 1.5mm and 15mm,
              start chain = going below,
                 V/.style = {circle, draw, 
                             fill=#1, 
                             node contents={}},
                            ]
            \node (n14) [V=yellow,on chain,label={left:$u_{1}$}];
            \node (n13) [V=blue,on chain,label={left:$u_{2}$}];
            \node (n12) [V=blue,on chain,label={left:$u_{3}$}];
            \node (n11) [V=blue,on chain,label={left:$u_{4}$}];
            \node (n10) [V=blue,on chain,label={left:$u_{5}$}];

            \node (n24) [V=red,right=1cm of n14,label={right:$v_{1}$}];
            \node (n23) [V=red,right=1cm of n13,label={right:$v_{2}$}];
            \node (n22) [V=red,right=1cm of n12,label={right:$v_{3}$}];
            \node (n21) [V=tgreen,right=1cm of n11,label={right:$v_{4}$}];
            \node (n20) [V=tgreen,right=1cm of n10,label={right:$v_{5}$}];
            \node [blue,fit=(n14) (n10),label=above:$\mathcal{U}$] {};
            \node [green,fit=(n24) (n20),label=above:$\mathcal{V}$] {};
            \draw[dashed]
            (n14) edge (n24)
            (n14) edge (n23)
            (n14) edge (n22)
            (n12) edge (n24)
            (n10) edge (n24)
            (n13) edge (n23)
            (n13) edge (n22);
            \draw[-]
            (n12) edge (n21)
            (n11) edge (n21) (n11) edge (n20)
            (n10) edge (n20);

            \end{tikzpicture}
            \caption{Remove Edges}
            \label{fig:example_edges}
    \end{subfigure}
    \hspace{10pt}
    \begin{subfigure}[]{0.2\textwidth}
        \centering
        \begin{tikzpicture}[
            node distance = 1.5mm and 15mm,
              start chain = going below,
                 V/.style = {circle, draw, 
                             fill=#1, 
                             node contents={}},
                            ]
            \node (n14) [V=yellow,on chain,label={left:$u_{1}$}];
            \node (n13) [V=blue,on chain,label={left:$u_{2}$}];
            \node (n12) [V=blue,on chain,label={left:$u_{3}$}];
            \node (n11) [V=blue,on chain,label={left:$u_{4}$}];
            \node (n10) [V=blue,on chain,label={left:$u_{5}$}];

            \node (n24) [V=white,right=1cm of n14,label={right:$v_{1}$}];
            \node (n23) [V=white,right=1cm of n13,label={right:$v_{2}$}];
            \node (n22) [V=white,right=1cm of n12,label={right:$v_{3}$}];
            \node (n21) [V=tgreen,right=1cm of n11,label={right:$v_{4}$}];
            \node (n20) [V=tgreen,right=1cm of n10,label={right:$v_{5}$}];
            \node [blue,fit=(n14) (n10),label=above:$\mathcal{U}$] {};
            \node [green,fit=(n24) (n20),label=above:$\mathcal{V}$] {};
            \draw[-]
            (n12) edge (n21)
            (n11) edge (n21)
            (n11) edge (n20)
            (n10) edge (n20);
        
            \end{tikzpicture}
            \caption{Remove Nodes}
            \label{fig:example_nodes}
    \end{subfigure}
    \caption{
    An example of a single iteration of \ours without the priority function. In this case, the degree of a sentence node serves as the priority score. Sentence nodes are in \textcolor{blue}{blue} and n-gram nodes in \textcolor{tgreen}{green}. The selected sentence node is \textcolor{tyellow}{yellow}, while connected n-gram nodes are \textcolor{red}{red}. Removed n-gram nodes are \textcolor{gray}{white}, with removed edges as dashed lines. Node $u_1$ is selected in the current iteration, and $u_4$ will be the next.
    }
    \label{fig:example}
\end{figure*}
\subsection{Modeling Datasets as Bipartite Graphs}
\label{sec:graph}

In our approach, we model the dataset as a bipartite graph to effectively represent the relationships between sentences and their constituent n-grams. A bipartite graph is a special type of graph whose vertices can be divided into two disjoint and independent sets such that every edge connects a vertex from one set to a vertex from the other set. Formally, a bipartite graph $\mathcal{G}=(\mathcal{U}, \mathcal{V}, \mathcal{E})$ consists of \textit{sentence nodes} ($\mathcal{U}=\{u_i\}^{N}_{i=1}$), \textit{n-gram nodes} ($\mathcal{V}=\{v_j\}^{M}_{j=1}$), and \textit{edges} ($\mathcal{E} \subseteq \mathcal{U} \times \mathcal{V}$). This structure allows us to capture the occurrence of n-grams within sentences, providing a foundation for selecting sentences that maximize n-gram coverage while adhering to specific priorities. We introduce the details of the priority for re-ranking the sentences based on both quality and diversity in \autoref{sec:priority}.

\paragraph{\ours}
Our objective is to select a subset of sentences, denoted as $\mathcal{S}$, from the entire dataset, constrained by a data selection budget $k$. The aim is to maximize the coverage of unique n-grams while aligning with a priority function $\phi(u)$ for each sentence $u \in \mathcal{U}$. As illustrated in \autoref{alg:graphfilter}, our method, referred to as \ours, operates iteratively by updating the graph structure to reflect the n-gram coverage as sentences are selected. The process begins with an empty set of selected sentences, $\mathcal{S} = \emptyset$, and a bipartite graph $\mathcal{G}$ that includes sentence nodes, n-gram nodes, and connecting edges. In each iteration, we select the sentence $u^{*} \in \mathcal{U}$ that has the highest priority score $\phi(u^{*})$, add $u^{*}$ to $\mathcal{S}$, and then remove $u^{*}$ from the set of remaining sentences $\mathcal{U}$. Next, we identify the n-grams covered by $u^{*}$, denoted as $\mathcal{V}_{u^*}$. We then remove all edges that connect $u^{*}$ to the n-gram nodes in $\mathcal{V}_{u^*}$. Subsequently, all edges connecting to nodes in $\mathcal{V}_{u^*}$ are eliminated from the graph. Note that the priority of each sentence $u \in \mathcal{U}$ is computed based on the most recent graph $\mathcal{G}$ during each iteration.

\paragraph{Set Cover Problem}
Our problem formulation is related to the classical \textit{set cover NP-hard problem} \citep{DBLP:books/fm/GareyJ79}. In the set cover problem, given a universe of elements and a collection of sets whose union comprises the universe, the objective is to identify the smallest number of sets whose union still contains all elements in the universe. Similarly, in a special case of our problem where the priority function assigns the same score to all sentences (i.e., $\phi(u) = 1$ for all $u \in \mathcal{U}$), and the goal is to find the minimal set of sentences that cover \textit{all} n-grams, our task becomes analogous to the set cover problem. In this scenario, the \textit{greedy approach} used in \autoref{alg:graphfilter} can be shown to have an approximation factor of $H(r)$ \citep{Vazirani:2001}, where $r$ is the maximum degree of the sentence nodes in the graph (the largest number of n-grams contained in any sentence), and $H(r) = \sum_{k=1}^r \frac{1}{k}$ is the $r$-th harmonic number. This relationship highlights the theoretical foundations of our method and provides insight into its performance guarantees in this special case.

\paragraph{A Minimalist Example}
Moreover, we present a minimalist example in \autoref{fig:example}. Initially, the bipartite graph is displayed in \autoref{fig:example_initial}. In \autoref{fig:example_selection}, the sentence node $u_1$ is selected as $u^{*}$ in \autoref{alg:graphfilter}, along with its associated n-gram nodes, $\mathcal{V}_{u^{1}}$, which are highlighted in red. \autoref{fig:example_edges} demonstrates the removal of edges connected to $u_1$ and $\mathcal{V}_{u^{1}}$, as indicated by dashed lines. Finally, \autoref{fig:example_nodes} illustrates the removal of isolated nodes, shown in white. The next selected sentence node is $u_4$. In this example, \ours can cover all the n-grams by selecting only $u_1$ and $u_4$.


By modeling the dataset as a bipartite graph and employing an iterative selection algorithm, \ours effectively selects a subset of sentences that maximizes n-gram coverage while adhering to specified priorities. \textit{Note that each SFT training instance comprises instructions and responses. In this work, we apply \ours solely to the instructions of the SFT data.}

\paragraph{Implementation}
In a brute-force implementation, the computational complexity of our algorithm is $\mathcal{O}(N)$ per iteration. This complexity results from the need to perform operations such as selecting the highest-priority sentence and removing edges, which involve scanning the sets of sentences ($\mathcal{U}$), n-grams ($\mathcal{V}$), and edges ($\mathcal{E}$). These sets are not optimized for efficient access or modification. To enhance computational efficiency, we employ a max-heap (or priority queue) to select the highest-priority sentence, allowing this selection to be performed in $\mathcal{O}(\log N)$ time per iteration. This reduces the selection complexity from $\mathcal{O}(N)$ to $\mathcal{O}(\log N)$. Additionally, the max-heap data structure facilitates the localization of priority updates to affected nodes, eliminating the need to enumerate all nodes and edges.

\subsection{Balancing Quality and Diversity with Priority Function}
\label{sec:priority}

As illustrated in \autoref{alg:graphfilter}, \ours naturally selects a subset with maximal n-gram coverage, emphasizing data diversity. However, the quality of the data is equally important for effective language model training. To balance both quality and diversity in our selection process, we define a priority function $\phi(u)$ for each sentence node $u \in \mathcal{U}$, which is used to re-rank the sentence nodes during selection.

\paragraph{Quality Metric}

For quality, we employ the \ifd as the quality measure \citep{li-etal-2024-superfiltering,li-etal-2024-quantity}. The \ifd metric evaluates the informativeness of a response by comparing the perplexity of the response conditioned on the instruction with the perplexity of the response alone. Formally, for a given sentence node $u$ associated with the instruction-response pair $(x, y)$, the quality priority metric is defined as:
\begin{equation}
\begin{aligned}
\model{Quality}(u) &= \ifd(x, y) = \frac{\textsc{ppl}(y \mid x)}{\textsc{ppl}(y)}, \\
\text{where } \textsc{ppl}(\vw) &= \exp\left( -\frac{1}{T} \sum_{t=1}^{T} \log P(w_t \mid \vw_{<t}) \right),
\label{eq:ifd}
\end{aligned}
\end{equation}
where $\textsc{ppl}(\vw)$ is the perplexity of the sentence $\vw$ with a length of $T$, $\textsc{ppl}(y)$ is the perplexity of the response $y$, and $\textsc{ppl}(y \mid x)$ is the perplexity of the response $y$ conditioned on the instruction $x$. A higher \ifd value indicates that the response is more relevant and informative given the instruction, thus reflecting higher quality. \textit{It is important to note that the choice of quality metric can be determined based on specific user needs and the quality scores can be precomputed prior to the selection process.}

\paragraph{Diversity Metric}

For diversity, we use the Term Frequency-Inverse Document Frequency (\model{TF-IDF}) as a measure of the significance of each n-gram within the dataset. The \model{TF-IDF} score of an n-gram $v$ is calculated as $\model{TF-IDF}(v) = \model{TF}(v) \times \model{IDF}(v)$, where $\model{TF}(v)$ (Term Frequency) is the number of times n-gram $v$ appears in the sentence, and $\model{IDF}(v)$ (Inverse Document Frequency) is defined as $\model{IDF}(v) = \log\left( \frac{N}{d_v} \right)$, with $N$ being the total number of sentences in the corpus, and $d_v$ being the number of sentences containing n-gram $v$. Furthermore, we compute the sum of \model{TF-IDF} scores of all n-grams (of varying lengths) present in the sentence:
\begin{equation} 
\label{eq:sum_tfidf} 
\model{Diversity}(u) = \sum_{v \in \mathcal{V}_u} \model{TF-IDF}(v),
\end{equation}
where $\mathcal{V}_u$ is the set of n-grams connected to sentence $u$ in the graph $\mathcal{G}$. \textit{In our work, $\mathcal{V}_u$ includes unigrams ($n=1$), bigrams ($n=2$), and trigrams ($n=3$) present in sentence $u$, capturing both word-level and phrase-level features.}

\paragraph{Priority Function}
To effectively prioritize sentences based on both quality and diversity, we combine the \model{Quality} score and the \model{Diversity} score for the sentence node $u$ into a single priority function:
\begin{equation}
\label{eq:priority_function} 
\phi(u) = \model{Quality}(u) \times \model{Diversity}(u).
\end{equation}
This function assigns higher priority to sentences that are both high-quality and contribute significantly to n-gram diversity. By integrating both quality and diversity into the priority function, our selection algorithm can effectively choose a subset of examples that are both high-quality and broadly representative of the entire dataset.

%

\section{Experiments}
\label{sec:experiments}

In this section, we initially outline our experimental setup in \autoref{sec:setup}, followed by our main results in \autoref{sec:main_results}.

\subsection{Experimental Setup}
\label{sec:setup}

\paragraph{Training Dataset}
\citet{DBLP:journals/corr/abs-2406-08464} utilize state-of-the-art open-source large language models (LLMs) to create a high-quality dataset collection known as \dataset{Magpie}. In our research, we employ the \dataset{Magpie} dataset, which is generated by \model{Llama-3-70B-Instruct} and comprises 300K training instances.\footnote{\url{https://huggingface.co/datasets/Magpie-Align/Magpie-Pro-300K-Filtered}} \textit{For this study, we choose a subset of 10K training instances using various selection methods from the entire dataset, unless otherwise stated.}

\paragraph{Baselines}
We compare our approach, \ours, with a diverse array of baseline methods:
\begin{itemize}
    \item \textbf{Heuristic}: (1) \random randomly selects a subset from the entire dataset; (2) \longest sorts the training instances in descending order based on the length of the instructions;
    \item \textbf{Quality-based}: (3) \ppl utilizes perplexity values, where larger values typically indicate higher difficulty and quality of training instances; (4) \armorm is the state-of-the-art open-sourced reward model presented by \citet{DBLP:journals/corr/abs-2406-12845}; 
    (5) \alpagasus demonstrates that state-of-the-art LLMs can be directly prompted for estimating data quality \citep{DBLP:conf/iclr/ChenLYWGYTS0HJ24}; (6) \deita leverages \model{ChatGPT} to synthesize a quality estimation dataset and fine-tune LLMs for data quality estimation \citep{DBLP:conf/iclr/0131Z00H24}; (7) \ifd indicates the Instruction-Following Difficulty (\model{IFD}) metric computed by smaller language models. \citet{li-etal-2024-quantity} introduce this method, while \citet{li-etal-2024-superfiltering} demonstrate that smaller models can be used for computing \model{IFD} scores.
    \item \textbf{Diversity-based}: (8) \kmeans clusters the training instances with the state-of-the-art sentence embedding model and selects the training instances that are closest to their respective cluster centroids \citep{DBLP:conf/soda/ArthurV07}; (9) \instag is designed for analyzing the SFT dataset by tagging the topics of training instances, and can be used for selecting the subset with the most diverse topics from the entire dataset \citep{DBLP:conf/iclr/LuY0LLTZZ24}.
\end{itemize}


We present more details of baselines in \autoref{appsec:baselines}. We conduct experiments using three diverse model backbones, including \gemma \citep{gemmateam2024gemma2improvingopen}, \mistral \citep{DBLP:journals/corr/abs-2310-06825}, and \llama \citep{DBLP:journals/corr/abs-2407-21783}. The optimization details are in \autoref{appsec:optimization}.

\begin{table*}[ht!]
\small
\centering
\setlength{\tabcolsep}{6pt}
\caption{
Main results given by \gemma, \mistral, and \llama on the standardized benchmarks and LLM-as-a-Judge benchmarks. \dataset{HS}, \dataset{G8K}, and \dataset{AE-2} correspond to \hellaswag, \gsmk, and \alpacaeval, respectively. The best results are highlighted in \textbf{bold}, and the second-best results are highlighted in \underline{underline}.
}
\begin{tabular}{lcccccccccccc}
\toprule
           & \multicolumn{5}{c}{Standardized}                                                     & \multicolumn{6}{c}{LLM-as-a-Judge}                                                                            & \multirow{3}{*}{\avgall} \\ \cmidrule(rl){2-6} \cmidrule(rl){7-12}
           & \mmlu          & \arc           & \dataset{HS}   & \dataset{G8K}   & \multirow{2}{*}{\avgbench} & \multicolumn{2}{c}{\dataset{AE-2}} & \multicolumn{3}{c}{\mtbench}                  & \multirow{2}{*}{\avgllm} &                          \\ \cmidrule(rl){7-8} \cmidrule(rl){9-11}
           & Acc            & Acc            & Acc            & Acc             &                            & LC               & WR              & \avgmt        & 1st           & 2nd           &                          &                          \\ \midrule
\multicolumn{13}{c}{\cellcolor{gray!30}\gemma}                                                                                                                                                                                                                             \\
\random    & 25.25          & 47.52          & 58.27          & \phantom{0}9.10 & 35.03                      & 10.77            & 13.73           & 4.73          & 5.44          & 4.01          & 29.01                    & 33.03                    \\
\longest   & 25.50          & 47.06          & 56.43          & \phantom{0}8.79 & 34.45                      & 10.40            & 13.10           & 4.79          & 5.54          & 4.05          & 29.17                    & 32.69                    \\
\ppl       & 23.34          & 47.58          & 59.04          & \phantom{0}6.48 & 34.11                      & 12.19            & 14.76           & \underline{4.98}    & \underline{5.75}    & 4.21          & 31.00                    & 33.07                    \\
\armorm    & 25.42          & \textbf{48.06} & 56.19          & \underline{10.62}     & 35.07                      & \textbf{13.40}   & \textbf{16.39}  & 4.84          & 5.55          & 4.14          & 30.92                    & 33.69                    \\
\alpagasus & 26.56          & 47.18          & 58.69          & 10.57           & 35.75                      & 13.12            & 15.76           & 4.89          & 5.68          & 4.11          & \underline{31.03}              & 34.18                    \\
\deita     & 28.72          & 47.51          & 58.35          & 10.16           & 36.18                      & 12.99            & 15.86           & 4.82          & 5.62          & 4.01          & 30.57                    & 34.31                    \\
\ifd       & \underline{28.82}    & 47.20          & 59.18          & \phantom{0}9.33 & 36.13                      & 12.87            & 15.55           & 4.88          & 5.49          & \underline{4.26}    & 30.81                    & \underline{34.36}              \\
\kmeans    & 28.39          & 46.96          & 56.59          & 10.31           & 35.56                      & 12.19            & 14.76           & \underline{4.98}    & 5.74          & 4.23          & 31.00                    & 34.04                    \\
\instag    & 27.60          & 47.75          & \textbf{59.98} & \phantom{0}9.86 & \underline{36.29}                & 12.75            & 15.47           & 4.79          & 5.45          & 4.13          & 30.31                    & 34.30                    \\ \midrule
\ours      & \textbf{29.06} & \underline{47.92}    & \underline{59.38}    & \textbf{10.71}  & \textbf{36.77}             & \underline{13.14}      & \underline{15.99}     & \textbf{5.01} & \textbf{5.77} & \textbf{4.25} & \textbf{31.64}           & \textbf{35.06}           \\ \midrule
\multicolumn{13}{c}{\cellcolor{gray!30}\mistral}                                                                                                                                                                                                                           \\
\random    & 25.50          & 52.17          & 67.44          & \phantom{0}9.17 & 38.57                      & 14.76            & 17.41           & 5.03          & 5.93          & 4.13          & 32.51                    & 36.55                    \\
\longest   & 25.17          & 52.11          & 67.32          & 10.30           & 38.73                      & 13.67            & 16.14           & 4.96          & 6.00          & 3.91          & 31.62                    & 36.36                    \\
\ppl       & 30.64          & 52.42          & \underline{69.31}    & \phantom{0}4.62 & 39.25                      & 13.60            & 16.18           & 4.98          & 6.01          & 3.95          & 31.70                    & 36.73                    \\
\armorm    & 28.84          & 50.85          & 68.85          & \phantom{0}9.63 & 39.54                      & \textbf{15.56}   & \textbf{18.89}  & 5.13          & 5.93          & 4.34    & \underline{33.43}              & 37.51                    \\
\alpagasus & 28.67          & 51.92          & 68.61          & \phantom{0}9.48 & 39.67                      & 14.67            & 18.14           & 5.21          & \underline{6.13}    & 4.30          & 33.40                    & 37.58                    \\
\deita     & 29.86          & 50.82          & 67.99          & 10.60           & 39.82                      & 14.08            & 16.49           & 5.03          & 5.93          & 4.13          & 32.18                    & 37.27                    \\
\ifd       & \textbf{33.59} & \underline{52.45}    & 68.56          & \phantom{0}9.93 & \underline{41.13}                & 13.59            & 16.75           & \underline{5.23}    & 6.01          & \underline{4.44}          & 32.92                    & \underline{38.40}              \\
\kmeans    & 28.77          & 50.58          & 67.81          & 11.55           & 39.68                      & 13.98            & 16.83           & 5.11          & 5.93          & 4.29          & 32.52                    & 37.29                    \\
\instag    & 28.29          & 50.99          & 67.44          & \textbf{12.59}  & 39.82                      & 14.55            & 17.36           & 5.11          & 5.86          & 4.36          & 32.84                    & 37.50                    \\ \midrule
\ours      & \underline{33.24}    & \textbf{52.48} & \textbf{69.69} & \underline{11.92}     & \textbf{41.83}             & \underline{15.16}      & \underline{18.85}     & \textbf{5.38} & \textbf{6.23} & \textbf{4.54} & \textbf{34.49}           & \textbf{39.38}           \\ \midrule
\multicolumn{13}{c}{\cellcolor{gray!30}\llama}                                                                                                                                                                                                                             \\
\random    & 49.55          & 52.00          & 67.30          & 22.14           & 47.75                      & 22.17            & 25.05           & 5.99          & 6.95          & 5.03          & 41.04                    & 45.51                    \\
\longest   & 44.52          & 50.56          & 67.99          & 24.56           & 46.91                      & 20.17            & 22.67           & 5.97          & 6.82          & 5.13          & 39.96                    & 44.59                    \\
\ppl       & 51.08          & 52.31          & \textbf{68.74} & 20.96           & 48.27                      & 20.38            & 22.87           & 6.02          & 7.02          & 5.01          & 40.28                    & 45.61                    \\
\armorm    & 47.84          & 52.24          & 68.11          & 24.64           & 48.21                      & \textbf{23.45}   & \underline{26.60}     & \underline{6.19}    & 7.14          & 5.24          & \underline{42.66}              & 46.36                    \\
\alpagasus & 49.90          & 51.63          & 68.40          & 25.89           & 48.96                      & 22.90            & 25.94           & 6.09          & 7.05          & 5.13          & 41.90                    & 46.60                    \\
\deita     & 48.49          & 52.40          & \underline{68.46}    & 25.78           & 48.78                      & 22.23            & 24.42           & 6.12          & 7.12          & 5.11          & 41.70                    & 46.42                    \\
\ifd       & 50.16          & 51.10          & 67.70          & \underline{27.45}     & 49.10                      & 22.54            & 24.68           & 6.13          & \textbf{7.23}    & 5.03          & 41.91                    & 46.70                    \\
\kmeans    & 51.98          & 51.35          & 67.15          & 25.12           & 48.90                      & 22.06            & 24.80           & 6.14          & 7.03          & \underline{5.25}    & 41.72                    & 46.51                    \\
\instag    & \underline{53.16}    & \underline{52.85}    & 67.86          & 25.85           & \underline{49.93}                & 22.10            & 24.64           & 6.13          & 7.05          & 5.21          & 41.72                    & \underline{47.19}              \\ \midrule
\ours      & \textbf{53.73} & \textbf{52.92} & 67.76          & \textbf{27.81}  & \textbf{50.55}             & \underline{22.95}      & \textbf{26.71}  & \textbf{6.26} & \underline{7.21} & \textbf{5.31} & \textbf{42.79}           & \textbf{47.97}          \\ \bottomrule
\end{tabular}

\label{tab:main}
\end{table*}

\paragraph{Evaluation}
We conduct evaluations on six popular benchmarks, categorized into two groups: 
\begin{itemize}
    \item \textbf{Standardized}: We assess the LLMs using \model{Lm-Evaluation-Harness} \citep{eval-harness} on four standardized benchmarks: \mmlu \citep{DBLP:conf/iclr/HendrycksBBZMSS21}, \arc \citep{DBLP:journals/corr/abs-1803-05457}, \hellaswag \citep{zellers-etal-2019-hellaswag}, and \gsmk \citep{DBLP:journals/corr/abs-2110-14168}. The model performance on these benchmarks is measured by accuracy. We use the macro-average accuracy across four benchmarks as the overall performance of this group, denoted as \avgbench. 
    \item \textbf{LLM-as-a-Judge}: We evaluate LLMs using \alpacaeval \citep{DBLP:journals/corr/abs-2404-04475} and \mtbench \citep{DBLP:journals/corr/abs-2306-05685}, with \model{gpt-4o-2024-05-13} as the judge. For \alpacaeval, \model{gpt-4-1106-preview} generates reference answers, and we report both the length-controlled win rate (LC) and the original win rate (WR). For \mtbench, performance is denoted as \avgmt, the macro-average across all categories. Overall performance of this group, \avgllm, is the macro-average of LC and \avgmt.
\end{itemize}
We define overall model performance, \avgall, as the macro-average of results from four standardized benchmarks, LC, and \avgmt. In calculating \avgall and \avgllm, \avgmt is scaled by $10 \times$ to align with a range of 1 to 100, matching other benchmarks. Further evaluation details are in \autoref{appsec:evaluation}.


\subsection{Main Results}
\label{sec:main_results}

\paragraph{\ours surpasses all baseline approaches.} 
As shown in \autoref{tab:main}, \ours consistently outperforms all baseline approaches across the three model backbones on both standardized benchmarks and LLM-as-a-Judge benchmarks. It achieves either the best or second-best results on most individual benchmarks. Specifically, in terms of \avgall, \ours outperforms the baselines by up to $+2.37$ for \gemma, $+3.02$ for \mistral, and $+3.38$ for \llama, compared to \longest. These results demonstrates the superiority of \ours which effectively combines the quality and diversity in selection.

\paragraph{Quality-based data selection approaches appear to exhibit biases towards specific benchmarks.}
Quality-based approaches often use neural models to estimate the quality of each training instance. However, these models display biases that can significantly affect downstream performance. As demonstrated in \autoref{tab:main}, models fine-tuned on subsets chosen by \armorm perform well on \alpacaeval but poorly on other benchmarks. Furthermore, the \ppl-selected subset consistently results in the worst performance on \gsmk, highlighting the risks of depending solely on neural models for selecting high-quality data.


\begin{table}[t]
\small
\centering
\setlength{\tabcolsep}{5pt}
\caption{Runtime (in hours) for selecting 10K training instances. $\dagger$ indicate the CPU-only method.}
\begin{tabular}{lc}
\toprule
                      & Runtime (hrs)   \\ \midrule
\ppl                  & \phantom{0}0.92 \\
\armorm               & \phantom{0}5.93 \\
\alpagasus            & 32.34           \\
\deita                & 22.65           \\
\ifd                  & \phantom{0}1.95 \\
\kmeans               & \phantom{0}2.26 \\
\instag               & 25.48           \\ \midrule
\ours                 & \phantom{0}2.48 \\
\quad w/o priority $\phi(u)$ & \phantom{0}0.53\rlap{\textsuperscript{$\dagger$}} \\ \bottomrule
\end{tabular}

\label{tab:compute}
\end{table}

\paragraph{\ours is highly efficient, with its variant running quickly on a CPU.}
Recent baselines typically rely on neural models for quality estimation, which generally require a GPU. We compare the runtimes of various baselines on a system equipped with an A100 80G GPU and 20 CPU cores, as shown in \autoref{tab:compute}. As elaborated in \autoref{sec:priority}, \ours defaults to using a quality estimation model for \(\model{Quality}(u)\). When utilizing \ifd, \ours completes its tasks in 2.48 hours, highlighting its efficiency. Notably, without using the priority function \(\phi(u)\) for re-ranking, \ours becomes even faster, taking only 0.53 hours on a CPU. This is up to $61\times$ faster than other baselines, compared to 32.34 hours by \alpagasus.




\section{Analysis}
\label{sec:analysis}

\begin{table}[t]
\small
\centering
\setlength{\tabcolsep}{5pt}
\caption{
Ablation study for n-gram combination with \llama. 
\checksymbol indicates that various n-grams are used.
}
\begin{tabular}{cccccc}
\toprule
\multicolumn{3}{c}{N-gram}           & \multirow{2}{*}{\avgbench} & \multirow{2}{*}{\avgllm} & \multirow{2}{*}{\avgall} \\ \cmidrule(rl){1-3}
Unigram    & Bigram     & Trigram    &                            &                          &                          \\ \midrule
\checksymbol & \checksymbol & \checksymbol & 50.55                      & 42.79                    & 47.97                    \\
\checksymbol &            &            & 49.02                      & 41.41                    & 46.48                    \\
           & \checksymbol &            & 49.09                      & 41.70                    & 46.63                    \\
           &            & \checksymbol & 49.84                      & 41.78                    & 47.15                    \\ \bottomrule
\end{tabular}

\label{tab:ngram}
\end{table}

\paragraph{Combining n-grams captures features at different levels.}
We examine the effectiveness of n-gram combinations, which are designed to capture both word-level and phrase-level features. The results are presented in \autoref{tab:ngram}. Our observations suggest that the variant of \ours, which integrates unigrams ($n=1$), bigrams ($n=2$), and trigrams ($n=3$), significantly outperforms other variations that do not incorporate n-gram combinations. Different n-grams capture features at varying levels, and merging them can effectively consolidate this information.

\begin{table}[t]
\small
\centering
\setlength{\tabcolsep}{5pt}
\caption{
Ablation study for $\model{Quality}(u)$ and $\model{Diversity}(u)$ in the priority with \llama.
\crosssymbol indicates the component is not used.
}
\begin{tabular}{ccccc}
\toprule
$\model{Quality}(u)$ & $\model{Diversity}(u)$ & \avgbench & \avgllm & \avgall \\ \midrule
\random              & \model{TF-IDF}         & 47.75     & 41.04   & 45.51   \\
\ifd                 & \model{TF-IDF}         & 50.55     & 42.79   & 47.97   \\
\ppl                 & \model{TF-IDF}         & 49.21     & 40.85   & 46.43   \\
\armorm              & \model{TF-IDF}         & 49.01     & 41.85   & 46.61   \\
\deita               & \model{TF-IDF}         & 49.11     & 41.97   & 46.73   \\
\crosssymbol         & \model{TF-IDF}         & 48.94     & 41.87   & 46.58   \\
\ifd                 & \crosssymbol           & 49.52     & 41.28   & 46.78   \\
\crosssymbol         & \crosssymbol           & 48.27     & 40.28   & 45.61   \\ 
\ifd & \model{MTLD} & 50.15	& 42.42 & 47.47 \\
\model{SkyworkRM} & \model{MTLD} & 49.51 & 42.01 & 46.93 \\
\bottomrule
\end{tabular}

\label{tab:priority}
\end{table}


\paragraph{Both $\model{Quality}(u)$ and $\model{Diversity}(u)$ in priority function enhance the data selection.}
We provide empirical evidence in \autoref{tab:priority} showcasing the effectiveness of our proposed priority function. By incorporating the $\model{Quality}(u)$ metric (using \ifd) and the $\model{Diversity}(u)$ metric (using \model{TF-IDF}) into \ours, we achieve superior performance across all evaluation metrics. This demonstrates that our combined priority function significantly enhances the model's ability to select high-quality and diverse training data. Omitting either the quality metric (\crosssymbol + \model{TF-IDF}) or the diversity metric (\ifd + \crosssymbol) results in noticeable performance declines. Furthermore, replacing the \ifd metric with \ppl as the quality measure leads to reduced performance, highlighting the importance of using optimal metrics. These findings support our decision to integrate quality and diversity in the priority.

\paragraph{\ours is compatible with various quality metrics.}
In this work, we leverage \ifd as the default quality metric due to its effectiveness and efficiency. However, \ours is designed to be flexible and is not limited to using \ifd alone. This flexibility allows the method to adapt to different evaluation needs by incorporating alternative quality metrics. As demonstrated in \autoref{tab:priority}, the $\model{Quality}(u)$ component of the priority function $\phi(u)$ can be replaced with various quality metrics, such as \ppl, \armorm, and \deita. This adaptability highlights the robustness and versatility of \ours.

\begin{table}[t]
\small
\centering
\setlength{\tabcolsep}{7pt}
\caption{Ablation study for the choice of n-gram order with \llama. \#nodes indicates the total number of nodes in the bipartite graph. RT indicates the runtime in hours.}
\begin{tabular}{cccccc}
\toprule
n-gram & \#nodes & RT (hrs)   & \avgbench & \avgllm & \avgall \\ \midrule
1      & 0.1M   & 2.12 & 49.02     & 41.41   & 46.48   \\
2      & 1.0M   & 2.30 & 49.58     & 42.14   & 47.31   \\
3      & 2.6M   & 2.48 & 50.55     & 42.79   & 47.97   \\
4      & 4.8M   & 3.38 & 50.11     & 42.63   & 47.43   \\
5      & 7.4M   & 4.58 & 50.44     & 42.81   & 47.95   \\ \bottomrule
\end{tabular}
\label{tab:trigram}
\end{table}

\pgfplotstableread[]{
x   y   label
23  9.6 \ours
46  11.3 Feb
60  12.8 Mar
54  9.8 Apr
}\advdata

\begin{figure*}[t]
\centering
\begin{subfigure}{0.33\textwidth}
  \centering
  
\begin{tikzpicture}
\begin{axis}[
    xlabel={Diversity $\uparrow$},
    ylabel={Quality $\uparrow$},
    x label style={at={(0.5,-0.1)}},
    y label style={at={(-0.1,0.5)}},
    xmin=70, xmax=110,
    ymin=25, ymax=45,
    width=5cm,
    height=5cm,
    grid=major,
    nodes near coords,
    every node near coord/.style={font=\tiny},
    every tick label/.append style={font=\tiny},
    scatter src=explicit symbolic
]

\addplot[
    scatter,
    only marks,
    color=green,
    mark=square*,
    visualization depends on={value \thisrow{text} \as \labela},
    nodes near coords={\labela}
]
table[meta=label] {
    x  y   label   text
    86.76  30.42  a  \random
    80.93  31.33  a  \longest
};

\addplot[
    scatter,
    only marks,
    color=blue,
    mark=triangle*,
    visualization depends on={value \thisrow{text} \as \labela},
    nodes near coords={\labela}
]
table[meta=label] {
    x  y   label   text
    76.88  29.34  b  \ppl
    75.12  40.77  b  \armorm
    84.09  30.31  b  \alpagasus
    87.08  30.31  b  \deita
    87.07  30.23  b  \ifd
};

\addplot[
    scatter,
    only marks,
    color=orange,
    mark=diamond*,
    visualization depends on={value \thisrow{text} \as \labela},
    nodes near coords={\labela}
]
table[meta=label] {
    x  y   label   text
    90.67  28.39  c  \kmeans
    93.45  30.65  c  \instag
};

\addplot[
    scatter,
    only marks,
    color=red,
    mark=*,
    visualization depends on={value \thisrow{text} \as \labela},
    nodes near coords={\labela}
]
table[meta=label] {
    x  y   label   text
    102.43  34.54  d  \ours
};

\end{axis}
\end{tikzpicture}
\caption{Quality-Diversity relationship}
\label{fig:tsne_qd_relation}
\end{subfigure}%
\begin{subfigure}{0.33\textwidth}
  \centering
  \includegraphics[scale=0.22]{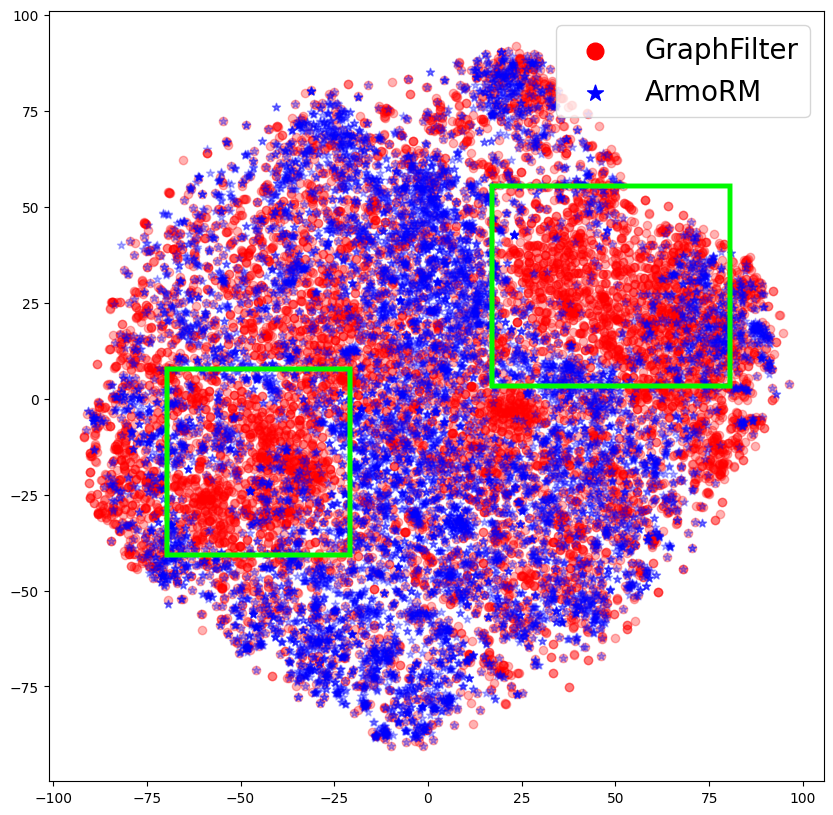}
  \caption{\ours vs. \armorm}
  \label{fig:tsne_armorm}
\end{subfigure}%
\begin{subfigure}{0.33\textwidth}
  \centering
  \includegraphics[scale=0.22]{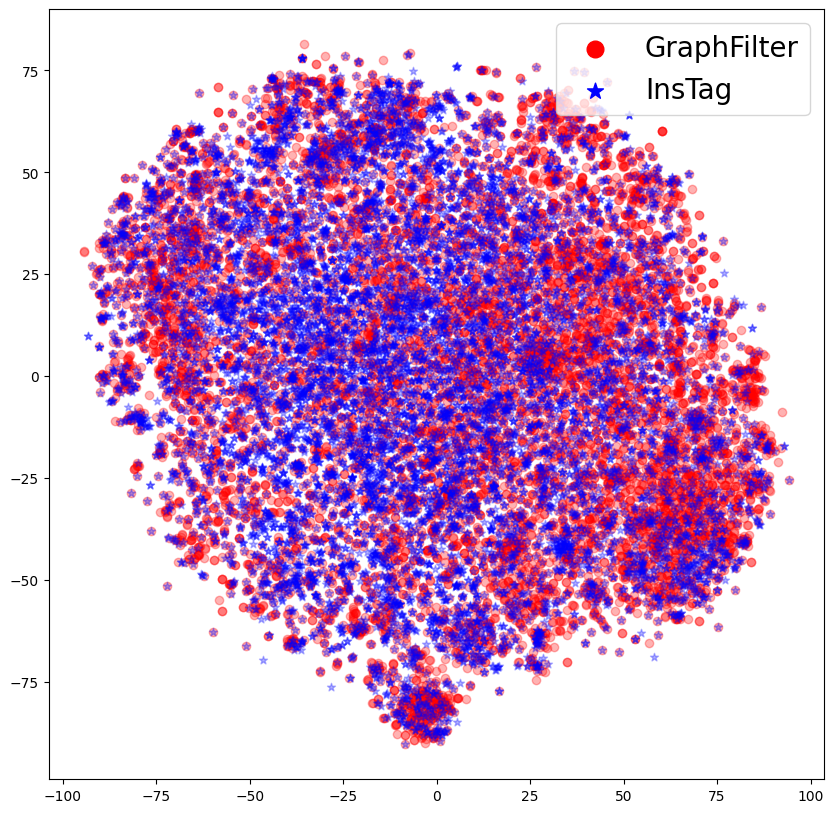}
  \caption{\ours vs. \instag}
  \label{fig:tsne_instag}
\end{subfigure}
\caption{
\autoref{fig:tsne_qd_relation} displays the quality-diversity relationships of subsets selected by different methods, with $\uparrow$ indicating a preference for higher values. \autoref{fig:tsne_armorm} shows the semantic diversity in a t-SNE plot of subsets from \ours and \armorm, where \textcolor{tgreen}{green rectangles} indicate data points chosen by \ours but not by \armorm. \autoref{fig:tsne_instag} depicts the semantic diversity in a t-SNE plot comparing subsets from \ours and \instag.
}
\label{fig:tsne}
\end{figure*}

\paragraph{The choice of trigrams ($n=3$) balances the model performance and efficiency.}
We conduct experiments with n-gram sizes from 1 to 5 using \llama. As shown in \autoref{tab:trigram}, our results indicate a significant performance improvement when moving from unigrams ($n=1$) to trigrams ($n=3$). However, beyond $n=3$, we observe diminishing or even negative returns. Furthermore, the number of n-gram nodes increases substantially with n (from 0.1M for unigrams to 7.4M for 5-grams), as well as the runtime (from 2.12 hours to 4.58 hours). These findings demonstrate that the trigrams ($n=3$) is the optimal choice for balancing performance and efficiency in our experiments.



\paragraph{\ours effectively balances quality and diversity in its selected datasets.}
In this section, we analyze the subsets selected by \ours and other methods, with results shown in \autoref{fig:tsne}. To confirm that \ours maintains quality and diversity, we measure lexical diversity using the \model{MTLD} metric \citep{mccarthy2010mtld} and assess data quality with the advanced reward model, \model{SkyworkRM} \citep{skyworkreward2024}.
As depicted in \autoref{fig:tsne_qd_relation}, \ours achieves the highest lexical diversity and ranks second in data quality. We also visualize \ours instructions compared with \armorm and \instag using the \model{bge-large-en-v1.5} model.
It is evident that \ours selects instructions not chosen by \armorm, shown by green rectangles in \autoref{fig:tsne_armorm}. Furthermore, \autoref{fig:tsne_instag} illustrates that \ours and \instag exhibit similar semantic diversity. These results suggest that \ours not only selects high-quality data but also maximizes dataset diversity.

\begin{table}[t]
\small
\centering
\setlength{\tabcolsep}{2pt}
\caption{
Applying \ours to instructions and responses with \llama. The \checksymbol indicates that \ours is applied. Lexical diversity is measured by \model{MTLD} \citep{mccarthy2010mtld}, and quality is assessed using \armorm, scaled by $100 \times$.
}
\begin{tabular}{cccccccc}
\toprule
\multicolumn{2}{c}{Content Type} & \multicolumn{3}{c}{Benchmarks} & \multicolumn{2}{c}{Lexical Diversity} & \multirow{2}{*}{Quality} \\ \cmidrule(rl){1-2} \cmidrule(rl){3-5} \cmidrule(rl){6-7} 
Inst.           & Resp.          & \avgbench  & \avgllm & \avgall & Inst.                   & Resp.       &                          \\ \midrule
 \checksymbol    &                & 50.55      & 42.79   & 47.97   & 102.43                  & 71.74       & 81.54                    \\
         & \checksymbol   & 47.16      & 39.71   & 44.68   & \phantom{0}90.22        & 73.57       & 81.52                    \\
\checksymbol    & \checksymbol   & 48.03      & 41.20   & 45.76   & \phantom{0}90.13        & 72.60       & 81.52                    \\ \bottomrule
\end{tabular}

\label{tab:sides}
\end{table}


\paragraph{Prioritizing instruction diversity most effectively improves model performance.}
Each SFT training instance comprises an instruction and its response. This study evaluates the impact of applying \ours to instructions, responses, or both on model performance. As shown in \autoref{tab:sides}, applying \ours only to instructions produces the best benchmark results, greatly improving lexical diversity in instructions with minimal effect on response diversity compared to other methods. Notably, all three variations maintain similar quality with different performances, underscoring the importance of instruction diversity.



\begin{figure}[t]
    \centering
    \begin{tikzpicture}[scale=0.5]
        \begin{axis}[
            width  = 0.85*\textwidth,
            height = 8cm,
            major x tick style = transparent,
            ybar=0,
            bar width=8pt,
            ymajorgrids = true,
            grid style={line width=.1pt, draw=gray!50},
            ylabel = {\LARGE $\Delta_{\textrm{\textsc{all}}}$},
            ylabel style={yshift=0cm},
            xlabel = {\LARGE Data selection budget},
            symbolic x coords={1K, 5K, 10K, 50K, 100K, 200K},
            xtick = data,
            scaled y ticks = false,
            enlarge x limits=0.15,
            enlarge y limits=0.15,
            ymax=2.5,
            legend cell align=left,
            legend columns=4,
            legend style={
                    at={(1,1)},
                    anchor=north east,
                    column sep=1ex
            }
        ]
    
            \addplot[style={orange,fill=orange}]
                    coordinates { (1K,2.18) (5K,1.83) (10K,1.08) (50K,0.11) (100K,-0.12) (200K,0.05) };

            \addplot[style={blue,fill=blue}]
                    coordinates { (1K,-0.08) (5K,0.58) (10K,1.02) (50K,0.78) (100K,0.38) (200K,0.25) };

            \addplot[style={red,fill=red}]
                    coordinates { (1K,1.58) (5K,1.68) (10K,2.03) (50K,1.53) (100K,0.58) (200K,0.33) };

            \legend{\ifd, \instag, \ours}


        \end{axis}
    \end{tikzpicture}
    
    \caption{
        Performance gap ($\Delta_{\textrm{\textsc{all}}}$) with respect to \avgall, comparing \ifd, \instag, and \ours against \random, across various data selection budgets.
    }
    \label{fig:data_scale}
\end{figure}

\paragraph{The priority of quality and diversity varies with data selection budgets, and \ours excels at balancing these two factors effectively.} 
After showcasing \ours's superiority, an open question remains: \textit{When should diversity be prioritized over quality, and vice versa?}
We hypothesize that the data selection budget plays a crucial role in determining the priority between quality and diversity and present the results in \autoref{fig:data_scale}. Our results indicate that the effectiveness of quality-based and diversity-based strategies is budget-dependent. Specifically, the quality-based \ifd excels with smaller budgets (1K and 5K instances), but its advantage diminishes as the budget increases. This suggests that quality-based methods with neural models may exhibit biases toward certain linguistic patterns, which limits model generalization when the budget is sufficiently large. Conversely, the diversity-based \instag performs poorly with small budgets but surpasses \ifd with larger ones. This observation demonstrates that diversity-based methods are more prone to introducing low-quality data with smaller budgets. Notably, \ours consistently achieves significant performance gains compared to \random across all budget levels. These findings show that the data selection budget influences the effectiveness of different approaches, and \ours successfully integrates both quality and diversity.

\section{Conclusion}

In this work, we formulate data selection as a set cover problem and introduce \ours, a novel method for data selection that models the dataset as a bipartite graph linking sentences to their constituent n-grams. To balance quality and diversity, we use a priority function that combines a quality metric with a diversity metric, allowing us to select subsets that enhance n-gram diversity and maintain high response quality. Our extensive experiments demonstrate \ours's effectiveness across three model backbones and six benchmark datasets. Compared to nine baseline methods, \ours consistently delivers superior model performance and computational efficiency. Our analyses validate our design choices, assess the subsets chosen by \ours and other methods, highlight the importance of instruction diversity, and examine the role of quality and diversity relative to subset sizes. We believe \ours lays the groundwork for more effective data selection strategies, encouraging further research in data selection for LLMs.

\section*{Impact Statement}
Our proposed method, \ours, aims to improve the efficiency and effectiveness of LLM training by enabling the selection of high-quality, diverse data subsets. This could lead to more resource-efficient model training processes and potentially better-performing models. While the primary impact of our work is methodological, we acknowledge that improvements in LLM training techniques could indirectly contribute to the broader adoption and application of these models across society. We believe the computational efficiency benefits of our approach align with efforts to reduce the environmental impact of training large models. As with any advancement in AI capabilities, we encourage thoughtful consideration of how improved language models might be deployed and used in practice.

\section*{Acknowledgements}
We would like to thank the anonymous reviewers and program chairs for their constructive feedback and valuable suggestions, which have greatly improved the quality of this paper. This work is partly supported by the ARC Future Fellowship FT190100039.

\bibliography{example_paper,iclr2025_conference,custom}
\bibliographystyle{icml2025}

\newpage
\appendix
\section{Experimental Setup}
\label{appsec:setup}

\subsection{Baselines}
\label{appsec:baselines}
In this work, we compare \ours against following baselines:
\begin{itemize}
    \item \textbf{\random} selects a random subset of size $k$ from the entire dataset, where $k$ is the designated data selection budget.
    \item \textbf{\longest} chooses the top-$k$ instances from the entire dataset, ranking them in descending order based on the number of words in each instruction.
    \item \textbf{\ppl} selects the top-$k$ instances from the entire dataset, sorted in descending order according to the perplexity values of the instructions. For the perplexity computation in this work, we utilize \model{gpt2} \citep{radford2019language}.\footnote{\url{https://huggingface.co/openai-community/gpt2}}
    \item \textbf{\armorm} represents one of the state-of-the-art reward models \citep{DBLP:journals/corr/abs-2406-12845}. It evaluates multiple rewards from diverse perspectives and integrates these rewards using a gating network.
    \item \textbf{\alpagasus} employs \model{gpt-3.5-turbo} to assess data quality \citep{DBLP:conf/iclr/ChenLYWGYTS0HJ24}. Given the improved model performance and limited budget, we substitute \model{gemma-2-27b-it} in this work, using the prompt illustrated in \autoref{fig:alpagasus_prompt}. \model{gemma-2-27b-it} is the state-of-the-art open large language model (LLM) and significantly surpasses \model{gpt-3.5-turbo} according to the Chatbot Arena Leaderboard.\footnote{\url{https://huggingface.co/spaces/lmsys/chatbot-arena-leaderboard}}
    \item \textbf{\deita} utilizes \model{ChatGPT} to create a quality estimation dataset and fine-tune large language models (LLMs) for evaluating data quality \citep{DBLP:conf/iclr/0131Z00H24}. We employ the official codes and models provided by \citet{DBLP:conf/iclr/0131Z00H24} for data selection.\footnote{\url{https://github.com/hkust-nlp/deita}}
    \item \textbf{\ifd} refers to the Instruction-Following Difficulty (\model{IFD}) metric, which is calculated using smaller language models. Introduced by \citet{li-etal-2024-quantity}, this method is shown by \citet{li-etal-2024-superfiltering} to provide \model{IFD} scores from smaller models that are as reliable as those from larger models. In this study, \model{gpt2} is used for computing these scores \citep{radford2019language}.
    \item \textbf{\kmeans} involves clustering training instances using a state-of-the-art sentence embedding model and selecting instances that are nearest to their respective cluster centroids \citep{DBLP:conf/soda/ArthurV07}. In this work, we begin by sampling 50K instances from the entire dataset and encoding their instructions into sentence embeddings using the \model{bge-large-en-v1.5} model.\footnote{\url{https://huggingface.co/BAAI/bge-large-en-v1.5}} These embeddings are used for training the \kmeans model with 10K clusters. Once the \kmeans model is established, we cluster the sentence embeddings of instructions for the entire dataset and select the instances closest to each cluster centroid.
    \item \textbf{\instag} is designed to analyze the SFT dataset by tagging the topics of training instances. It can be used to select a subset with the most diverse topics from the entire dataset \citep{DBLP:conf/iclr/LuY0LLTZZ24}. We utilize the official codes and models released by \citet{DBLP:conf/iclr/LuY0LLTZZ24} for data selection.\footnote{\url{https://github.com/OFA-Sys/InsTag}}
\end{itemize}

\begin{figure}[t]
    \centering
    \footnotesize
    \begin{Verbatim}[frame=single, fontsize=\tiny, breaklines=true, breakanywhere=true]
### System: 
We would like to request your feedback on the performance of AI assistant in response to the instruction and the given input displayed following.

###Instruction:
{instruction}

### Input:
{input}

### Response:
{output}

### USER:
Please rate according to the accuracy of the response to the instruction and the input. Each assistant receives a score on a scale of 0 to 5, where a higher score indicates higher level of the accuracy. Please first output a single line containing value indicating the scores. In the subsequent line, please provide a comprehensive explanation of your evaluation, avoiding any potential bias.
    \end{Verbatim}
    \caption{The prompt used for \alpagasus annotation.}
    \label{fig:alpagasus_prompt}
\end{figure}

\subsection{Optimization}
\label{appsec:optimization}

\paragraph{Hyperparameters}
In this study, all experiments utilize the same set of hyperparameters. Specifically, we employ a batch size of 64, a learning rate of $2 \times 10^{-5}$, a warmup ratio of 0.05, and a linear learning rate schedule. All the experiments run for 3 epochs.

\paragraph{Computation Infrastructure}
For this study, all methods are trained using two A100 80GB GPUs, which are interconnected via PCIe.

\subsection{Evaluation}
\label{appsec:evaluation}

In this work, we evaluate the approaches on six widely used benchmarks:

\begin{itemize}
    \item \textbf{\mmlu} \citep{DBLP:conf/iclr/HendrycksBBZMSS21} is a benchmark designed to assess knowledge acquired during pretraining, by evaluating models exclusively in zero-shot and few-shot settings. It covers 57 subjects across STEM, the humanities, social sciences, and more, totaling approximately 14,000 test examples.
    
    \item \textbf{\arc} \citep{DBLP:journals/corr/abs-1803-05457} is a multiple-choice question-answering dataset containing questions from science exams for grades 3 to 9, amounting to approximately 4,000 test examples.
    
    \item \textbf{\hellaswag} \citep{zellers-etal-2019-hellaswag} is a challenging dataset for evaluating commonsense natural language inference, which is particularly difficult for state-of-the-art models, though its questions are trivial for humans. It contains approximately 10,000 test examples.
    
    \item \textbf{\gsmk} \citep{DBLP:journals/corr/abs-2110-14168} comprises a collection of diverse grade school math word problems created by human problem writers, containing approximately 1,000 test examples.
    
    \item \textbf{\alpacaeval} \citep{DBLP:journals/corr/abs-2404-04475} is an automated tool for evaluating instruction-following language models. Its test set consists of 805 instructions generated by large language models (LLMs). Models are evaluated based on the winning rate against a reference answer, judged by a state-of-the-art LLM, such as \model{gpt-4}. \alpacaeval is an upgraded version of the original \dataset{AlpacaEval}, featuring reduced length bias for a fairer evaluation of responses of varying lengths.
    
    \item \textbf{\mtbench} \citep{DBLP:journals/corr/abs-2306-05685} is a multi-turn test set containing 80 questions that cover 8 aspects: writing, roleplay, reasoning, math, coding, extraction, STEM, and humanities. A state-of-the-art LLM, such as \model{gpt-4}, is used to score model outputs on a scale from 1 to 10.
\end{itemize}

\end{document}
